\documentclass[a4paper,fleqn]{cas-dc}

\usepackage[authoryear]{natbib}
\usepackage{amsmath}

\usepackage{adjustbox}
\usepackage{mathtools}
\usepackage{comment}
\usepackage{hyperref}
\usepackage{threeparttable}
\usepackage[T1]{fontenc}
\usepackage{booktabs}
\usepackage{array}
\usepackage{amsmath}
\usepackage{graphicx,verbatim}
\usepackage{url}
\usepackage{hyperref}
\usepackage{multirow}
\usepackage{amssymb}

\def\tsc#1{\csdef{#1}{\textsc{\lowercase{#1}}\xspace}}
\tsc{WGM}
\tsc{QE}

\begin{document}
\let\WriteBookmarks\relax
\def\floatpagepagefraction{1}
\def\textpagefraction{.001}

\shorttitle{LV-Net: Anatomy-aware lateral ventricle shape modeling}    

\shortauthors{W. Park}  

\title [mode = title]{LV-Net: Anatomy-aware lateral ventricle shape modeling with a case study on Alzheimer's disease}  


\author[inst1]{Wonjung Park}
\author[inst1]{Suhyun Ahn}
\author[inst1]{Jinah Park}
\author{for the Alzheimer’s Disease Neuroimaging Initiative\footnote{%
Data used in preparation of this article were obtained from the Alzheimer’s Disease Neuroimaging Initiative (ADNI) database (adni.loni.usc.edu). 
As such, the investigators within the ADNI contributed to the design and implementation of ADNI and/or provided data but did not participate in analysis or writing of this report. 
A complete listing of ADNI investigators can be found at: \url{http://adni.loni.usc.edu/wp-content/uploads/how_to_apply/ADNI_Acknowledgement_List.pdf}}}
\author{the Australian Imaging Biomarkers and Lifestyle flagship study of ageing\footnote{Data used in the preparation of this article was obtained from the Australian Imaging Biomarkers and Lifestyle flagship study of ageing (AIBL) funded by the Commonwealth Scientific and Industrial Research Organisation (CSIRO) which was made available at the ADNI database (www.loni.usc.edu/ADNI). The AIBL researchers contributed data but did not participate in analysis or writing of this report. AIBL researchers are listed at www.aibl.csiro.au.}}



\affiliation[inst1]{organization={School of Computing, KAIST},
            addressline={291 Daehak-ro, Yuseong-gu}, 
            city={Daejeon},
            postcode={34141}, 
            country={Republic of Korea}}




\begin{abstract}
Lateral ventricle (LV) shape analysis holds promise as a biomarker for neurological diseases; however, challenges remain due to substantial shape variability across individuals and segmentation difficulties arising from limited MRI resolution. We introduce LV-Net, a novel framework for producing individualized 3D LV meshes from brain MRI by deforming an anatomy-aware joint LV-hippocampus template mesh. By incorporating anatomical relationships embedded within the joint template, LV-Net reduces boundary segmentation artifacts and improves reconstruction robustness.
In addition, by classifying the vertices of the template mesh based on their anatomical adjacency, our method enhances point correspondence across subjects, leading to more accurate LV shape statistics. We demonstrate that LV-Net achieves superior reconstruction accuracy, even in the presence of segmentation imperfections, and delivers more reliable shape descriptors across diverse datasets. Finally, we apply LV-Net to Alzheimer's disease analysis, identifying LV subregions that show significantly associations with the disease relative to cognitively normal controls.
The codes for LV shape modeling are available at \url{https://github.com/PWonjung/LV_Shape_Modeling}.
\end{abstract}


\begin{highlights}
\item We propose an automated framework for reconstructing individual lateral ventricle shapes from brain structural MRIs. 
\item We introduce an anatomy-aware joint template mesh that supports robust and anatomically accurate modeling of lateral ventricle shapes.
\item We analyze shape differences in the lateral ventricles between patients with Alzheimer's disease and cognitively normal controls.
\end{highlights}



\maketitle

\section{Introduction}\label{section1}
The lateral ventricle (LV), a C-shaped cavity located at the center of the brain, serves as a valuable indicator of overall brain morphology. Numerous LV shape analysis studies \citep{dong2020applying,johnson2013hippocampal,styner2005morphometric,trimarchi2013mri} have demonstrated that the LV undergoes distinct deformations under both normal and pathological conditions.
However, traditional employed methods such as SPHARM-PDM~\citep{spharm}, LDDMM~\citep{beg2005computing}, and ShapeWorks\citep{cates2017shapeworks}, not specifically designed for the LV, often exhibit limitations in robustness and accuracy when capturing its complex morphological variations.
Furthermore, most previous research~\citep{styner2005morphometric, lee2013morphometric} has conducted analysis only on the partial LV (i.e., the superior) since these methods are not suitable for disconnected and noisy segmented masks of inferior LV. 
Therefore, a robust modeling method tailored to the entire LV is necessary.

For accurate LV shape modeling, three major challenges must be addressed simultaneously: (1) significant shape variation, (2) segmentation errors, and (3) difficulty in establishing anatomically consistent point correspondence. First, substantial inter-subject variability requires robust shape modeling methods. For instance, some individuals have prominent occipital horns, while others lack this structure entirely. This high variability complicates the modeling of morphological variation.
Second, erroneous LV segmentation should be corrected using prior LV anatomy knowledge to model accurate shapes. Despite the continuous nature of the LV, many segmented LV masks contain holes and disconnected regions due to the limited MRI resolution. These boundary segmentation errors are particularly prominent in the inferior LV, where the structure is thin and thus challenging to delineate accurately, given its proximity to the hippocampus.
Finally, establishing anatomically consistent point correspondences is crucial for precise clinical analysis on statistical shapes. Given that the LV boundary is defined by surrounding subcortical structures such as the thalamus, caudate, and hippocampus, corresponding points must align with the same anatomical regions across subjects \citep{ferrarini2008mmse, park2024lateral}. However, traditional methods, relying solely on LV geometry, can lead to inaccurate correspondences and potentially misleading clinical interpretations.

To address these challenges, we introduce LV-Net, an automatic framework for reconstructing individual 3D LV shapes from brain MRI using an anatomy-aware deformable joint template mesh.
In the LV-Net framework, we propose a novel LV-hippocampus joint template mesh, which facilitates the accurate reconstruction of the entire LV including its {\it inferior} region by leveraging anatomical relationships with the hippocampus. By incorporating this prior knowledge, LV-Net can robustly model LV shapes even in the presence of boundary segmentation errors.
Furthermore, to ensure anatomically consistent point correspondences across subjects, we embed periventricular region information into the template mesh vertices. Leveraging this information, LV-Net deforms vertices to align with identical peripheral regions, preserving anatomical consistency.

In our experiments, we thoroughly evaluate LV-Net against conventional shape modeling methods used for LV shape analysis. Quantitative results demonstrate that LV-Net achieves superior shape alignment and produces more accurate shape statistics across diverse datasets. 
Furthermore, comprehensive studies confirm that the anatomy-aware joint template mesh effectively mitigates the impact of segmentation errors, particularly in the inferior LV. In addition, we validate that the embedded peripheral information of the template mesh improves point correspondences, accurately reflecting adjacent brain structures.
Finally, we demonstrate the clinical utility of LV-Net in Alzheimer's disease analysis, identifying LV subregions significantly associated with the disease compared to cognitively normal controls.
We expect our LV-tailored modeling framework will pave the way for precise local and global analysis of LV shape, providing a valuable foundation for various neurological diseases.

\section{Related work}
\subsection{Lateral ventricle shape modeling}
As a cavity located at the center of the brain and surrounded by subcortical structures, the lateral ventricle (LV) serves as a valuable proxy for the global structure of the brain anatomy. Numerous studies have investigated the volume and shape of the LV in relation to various pathologies, including normal aging~\citep{trimarchi2013mri, bethlehem2022brain}, Alzheimer’s disease~\citep{ferrarini2006shape, tang2014shape}, Parkinson’s disease~\citep{apostolova2010hippocampal}, and other neurological conditions~\citep{styner2005morphometric, richards2020increased}.

For LV shape analysis, spherical harmonic~\citep{spharm} and diffeomorphic deformation-based~\citep{beg2005computing, lddmm, dong2020applying} approaches have been widely adopted. These methods typically reconstruct individual shapes under the assumption that the input binary masks are fully connected. However, in practice, LV masks often include missing or incomplete regions, particularly in the inferior horn. Given this challenges, spherical harmonic-based studies have focused their analysis on the superior portion of the LV without inferior horn~\citep{lee2013morphometric, kang2018lateral}.

To address the issue of fragmented binary masks, point cloud-based methods~\citep{cates2017shapeworks} can be employed, which directly infer correspondences from the boundary points of segmented regions. However, due to the limited resolution of MRI scans, these boundary point clouds capture anatomically inaccurate structures, making it difficult to reconstruct biologically plausible LV shapes. This limitation becomes particularly critical given the strong morphological interdependence between the inferior LV and the hippocampus~\citep{apostolova2012hippocampal, oliveira2019normal}. Therefore, comprehensive modeling of the entire LV is essential for a more accurate understanding of brain structures.

Furthermore, conventional approaches generally extract LV masks from MRI scans without integrating adjacent subcortical structures, potentially leading to anatomically inconsistent correspondences across individuals. One prior study that examined associations between Mini-Mental State Examination (MMSE) scores and LV morphology attempted to improve inter-subject correspondence by manually adjusting landmark points to align with neighboring regions~\citep{ferrarini2008mmse}. Although such manual refinement can enhance anatomical accuracy, it arises subjectivity and scalability issues. These limitations highlight the need for automated methods that ensure biologically meaningful correspondences particularly in periventricular regions.

\subsection{AI-based 3D shape reconstruction}
With advancements in deep learning, several works have proposed reconstruction of 3D shapes in mesh representation from other formats such as point clouds~\citep{Hanocka2020p2m} and voxels~\citep{wickramasinghe2020voxel2mesh}.
In the medical image analysis field, since MRI and its segmentation maps are typically in voxel representation, many studies have attempted 3D mesh reconstruction employing deep learning architectures~\citep{bastian2023s3m, el2024universal}.

Recent research has introduced general deep learning methods for shape reconstruction applicable across various anatomical structures~\citep{bastian2023s3m, iyer2023mesh2ssm, el2024universal, park2025ai}.
Moreover, to reconstruct biologically plausible shapes given the specific challenges posed by target organs, several studies have proposed reconstruction techniques for vertebrae~\citep{kim2024vertebral}, the heart~\citep{kong2021deep, ye2023neural}, and cortical surfaces~\citep{cruz2021deepcsr, ma2022cortexode}. These approaches aim to recover surface meshes that maintain anatomical correctness and facilitate shape-based statistical analyses by offering the shape correspondences across shapes.

Considering the anatomical complexity of LV, a cavity surrounded by subcortical structures, tailored modeling approaches are required.
For LV shape reconstruction, to the best of our knowledge, our previous work PeriNet~\citep{park2024lateral} was the first attempt to reconstruct LV shapes using deep learning architectures. This approach incorporates periventricular anatomical information to establish anatomically meaningful correspondences for longitudinal shape deformation analysis. PeriNet reconstructs meshes via the marching cubes algorithm, without relying on template meshes or predefined point correspondences. Building upon this idea, we extend the framework by introducing a deformable template mesh, which enables explicit modeling of shape correspondences and more detailed statistical analysis of shape variations.

\section{Method}
In this section, we describe the anatomy-aware deformable joint template mesh and the LV-Net pipeline for reconstructing LV shapes.

\begin{figure*}[t]
\includegraphics[width=\textwidth]{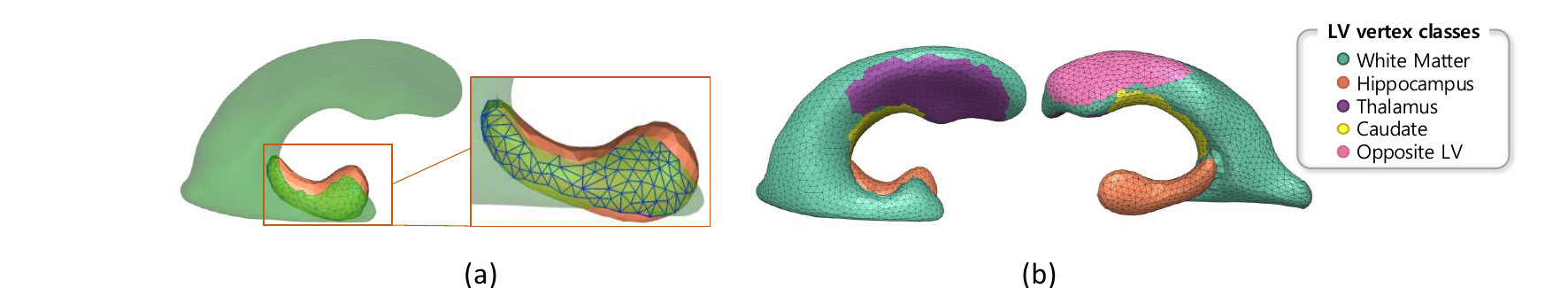}\centering
\caption{\textbf (a) LV-hippocampus joint template mesh with shared vertices between LV and hippocampus and (b) anatomy-aware joint template mesh with adjacent area information.} \label{fig:temp_mesh}
\end{figure*}

\subsection{Anatomy-aware deformable joint template mesh}
The anatomy-aware deformable joint template mesh serves as the initial template for LV-Net (see \autoref{fig:temp_mesh}). For robust and accurate shape modeling, this template consists of two key components: (1) the LV-hippocampus joint mesh and (2) peripheral region information embedded in the LV mesh vertices.


First, to preserve anatomical correctness, we introduce the LV-hippocampus joint template mesh of \autoref{fig:temp_mesh} (a), which has shared vertices $V^{shared}$ along the boundary between the LV and hippocampus. These vertices enforce structural coherence, ensuring that even when inferior LV masks are incomplete or missing, LV maintains the structure along the hippocampus boundary.
The LV mesh ${M}^{L}$ comprises vertices green-colored ${V}^{L}$ along with ${V}^{shared}$, while the hippocampus mesh ${M}^{h}$ consists of orange-colored vertices ${V}^{h}$ and ${V}^{shared}$.
During iterative reconstruction, the positions of all vertices $V=\left\{{V}^{L},{V}^{h},{V}^{shared}\right\}$ are simultaneously optimized to closely match the target shape. 

For anatomically consistent point correspondences across subjects, we embed peripheral region information into the template mesh vertices (see \autoref{fig:temp_mesh} (b)). 
Specifically, each LV vertex is classified as one of five adjacent brain structures: white matter, hippocampus, thalamus, caudate, or the opposite LV.
During shape deformation, these embedded anatomical labels guide the vertices to align with the same subcortical structures across subjects. This guarantees accurate correspondence, facilitating reliable shape analysis.

In this work, we define the template mesh with an edge length of approximately 2 mm, and the LV mesh $M^L$ consists of 2,490 vertices including 96 shared vertices $V^{shared}$.

\subsection{Optimization from template mesh to target shape}
We propose an automatic LV shape modeling framework from brain MRIs. The process consists of two key stages: generating the target point cloud from a brain MRI segmentation map and iteratively optimizing the deformable template mesh to match the target shape.
\begin{figure*}[!t]\centering
\includegraphics[width=\textwidth]{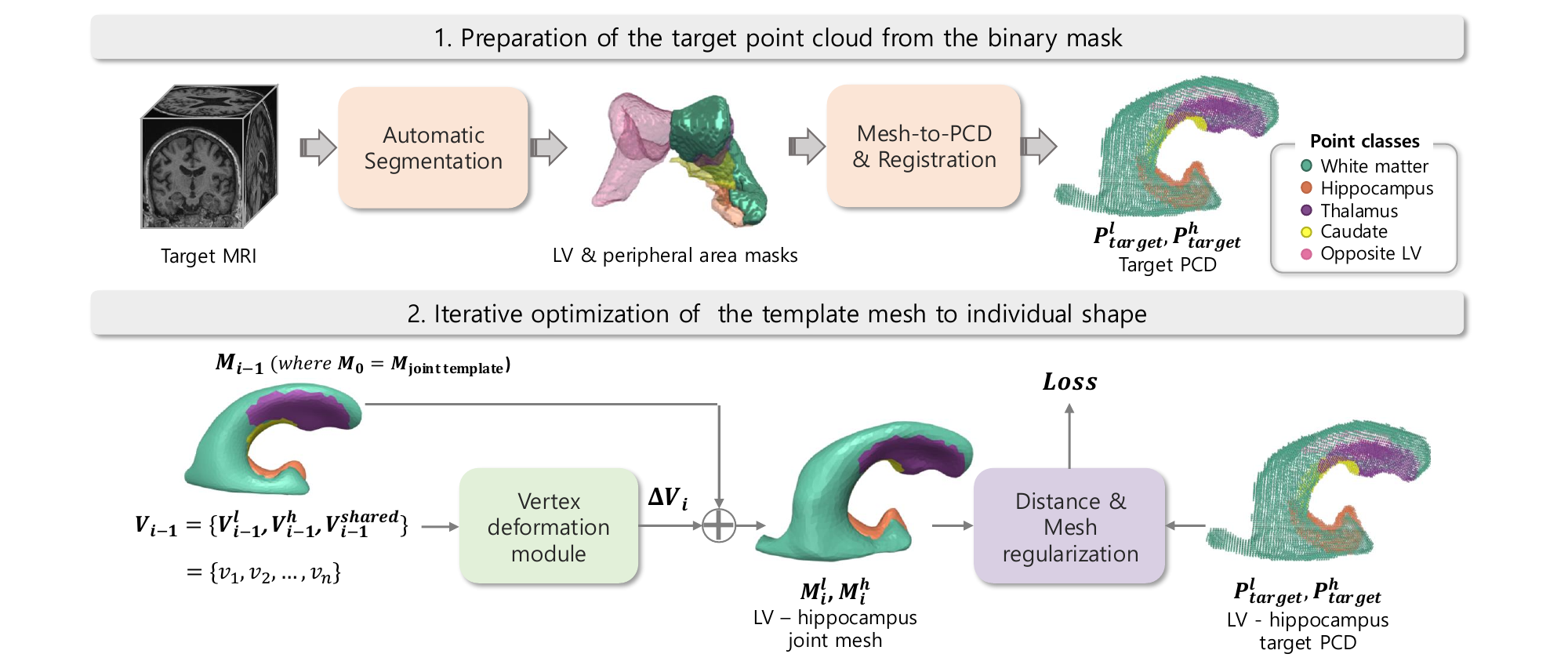}
\caption{Overall framework to model the target LV and hippocampus mesh from the joint template mesh.}\label{fig:pipeline}
\end{figure*}

\subsubsection{Preparation of the target point cloud}
To construct the target shape, we generate a point cloud from the segmentation map of an MRI. 
In this study, to obtain segmentation maps, we employ SynthSeg~\citep{billot2023synthseg}, a widely used automatic MRI segmentation model.
We extract boundary points from the segmented LV and hippocampus masks to form the target point clouds, ${P}_{target}^{L}$ for LV and ${P}_{target}^{H}$ for hippocampus, respectively.
To ensure anatomical consistency, we classify each point in ${P}_{target}^{L}$ to five peripheral subcortical structures (see \autoref{fig:pipeline}) by searching its $2\times2\times2$ neighborhood voxels.
Then, the target point cloud is aligned with the joint template mesh space using a two-step registration process. We first perform a global alignment (\textit{i.e.}, rigid transform) via Iterative Closest Point~\citep{rusinkiewicz2001efficient} using the combined left and right LV.
Afterward, scale adjustment is applied to the template mesh along x,y, and z dimensions to have the smallest distance to the target point cloud.

\subsubsection{Iterative optimization of the deformable template mesh}
To reconstruct the LV shape from the target point cloud, we iteratively optimize the joint template mesh by deforming its vertices to align with the given target (see \autoref{fig:pipeline}). This process ensures that the reconstructed LV maintains anatomical accuracy while deforming to individual shapes. At each iteration, a PointNet-based vertex deformation module~\citep{PointNet} predicts vertex displacements, and then the mesh progressively refines its shape to match the target LV structure.

The optimization is guided by a loss function comprising two main losses: the distance loss \( \mathcal{L}_{\text{dist}} \) and regularization loss \( \mathcal{L}_{\text{reg}} \).
The distance loss minimizes the discrepancy between the deformed mesh and the target point cloud, while the regularization loss ensures smooth and realistic deformations.
The distance loss is further decomposed into three terms:
\( \mathcal{L}_{\text{dist}}^{L} \) and \( \mathcal{L}_{\text{dist}}^{H} \), which aligns the LV and hippocampus mesh with the target corresponding point cloud ${P}_{target}^{L}$ and ${P}_{target}^{H}$,
and \( \mathcal{L}_{\text{dist}}^{\text{peri}} \), which enforces anatomical correspondence by guiding vertices toward their predefined peripheral regions. The overall distance loss is:
\begin{equation}\label{equ:dist}
\mathcal{L}_{dist} = \mathcal{L}_{dist}^{L}+\mathcal\mathcal{L}_{dist}^{H}+\sum_{p=1}^{m}\mathcal{L}_{dist}^{\text{peri}_{p}}
\end{equation}
\begin{align*}
\text{where}\;\; \mathcal{L}_{dist}^{i} = {\lambda}_{cf}\mathcal{L}_{cf}^{i}+{\lambda}_{pm}\mathcal{L}_{pm}^{i}+{\lambda}_{mp}\mathcal{L}_{mp}^{i} \\
(i=\{L,H,{peri}_{p}\}).\\
\end{align*}

\noindent where $\mathcal{L}_{cf}$ is the Chamfer distance between the deformed mesh ${M}_{i}$ and the point cloud ${P}_{target}^{i}$. $\mathcal{L}_{pm}$ measures the point-to-face distance from ${P}_{target}^{i}$ to ${M}_{i}$, while $\mathcal{L}_{mp}$ is vice-versa. In our framework, we define four peripheral regions: white matter, the thalamus, caudate, and opposite LV (\textit{i.e.}, \( m = 4 \) in \autoref{equ:dist}). 
The hyperparameters for {${\lambda}_{{cf}},{\lambda}_{{pm}}$, ${\lambda}_{{mp}}$} are {2, 1.4, 0.6}. A higher weight is assigned to $\mathcal{L}_{pm}$ compared to $\mathcal{L}_{mp}$, accounting for missing points in the target point cloud due to segmentation errors.

The regularization loss ensures plausible mesh deformation and is derived from each deformed LV and hippocampus meshes.
The regularization loss is :
\begin{equation}
\mathcal{L}_{reg} = \mathcal{L}_{reg}^{L}+\mathcal{L}_{reg}^{h}
\end{equation}
\begin{flalign*}
\text{where} \;\; \mathcal{L}_{reg}^{i} = 
& {\lambda}_{vert}||\Delta vert^{i}||_2 +
  {\lambda}_{norm}||\Delta norm^{i}||_2 \\
& + {\lambda}_{edge}\mathcal{L}_{edge}^{i}
  + {\lambda}_{cons}\mathcal{L}_{cons}(norm^{i})\\
& + {\lambda}_{lap}\mathcal{L}_{lap}^{i}, \quad (i=\{L,H\})
\end{flalign*}
\noindent where \(\Delta vert \) represents the distance of the vertices moved, which encourages minimal vertex displacement as the deformable mesh approaches the target shape. The second term for \(\Delta norm\) penalizes deviations of the normal vectors from the template mesh normals, aiding in establishing more precise point correspondences.  
\( \mathcal{L}_{\text{edge}} \) is the variance of edge lengths, which prevents the formation of a skewed mesh. Normal consistency \( \mathcal{L}_{\text{cons}} \) and Laplacian smoothness with cotangent curvature weight~\citep{desbrun1999implicit} \( \mathcal{L}_{\text{lap}} \) are employed to ensure a smooth deformed mesh. 
The hyperparameters for {\( \lambda_{\text{vert}}, \lambda_{\text{norm}}, \lambda_{\text{edge}}, \lambda_{\text{cons}} \), and \( \lambda_{\text{lap}} \)} are set to {1, 1, 1000, 100, 300}.
We used the AdamW optimizer~\citep{loshchilov2017decoupled_adamw} with an initial learning rate of \( 5 \times 10^{-4} \), halved every 1000 iterations over 5000 iterations.

\section{Experiments and Results}

\subsection{Experimental setup}

\textbf{Dataset.}
To validate our LV-Net, we used three publicly available brain MRI datasets: OASIS~\citep{OASIS}, ADNI~\citep{jack2008alzheimer}, and AIBL~\citep{ellis2014rates}. We performed shape reconstruction on both cognitively normal and demented groups within each dataset to demonstrate the generalizability of LV-Net across highly variable LV morphologies. 

For ADNI dataset, the data used in the preparation of this article were obtained from the Alzheimer’s Disease Neuroimaging Initiative (ADNI) database (adni.loni.usc.edu). The ADNI was launched in 2003 as a public-private partnership, led by Principal Investigator Michael W. Weiner, MD. The primary goal of ADNI has been to test whether serial magnetic resonance imaging (MRI), positron emission tomography (PET), other biological markers, and clinical and neuropsychological assessment can be combined to measure the progression of mild cognitive impairment (MCI) and early Alzheimer’s disease (AD). For AIBL dataset, the data was collected by the AIBL study group. AIBL study methodology has been reported previously~\citep{ellis2009australian}.

\textbf{Baselines.}
To demonstrate the effectiveness of the tailored design of LV-Net, we compared our method to prior shape modeling methods, including LDDMM~\citep{lddmm} and ShapeWorks~\citep{cates2017shapeworks}. SPHARM-PDM~\citep{spharm} was excluded, as it requires connected binary masks, which are often impractical for the inferior LV.
LDDMM is a registration-based method that deforms a template mesh using diffeomorphic transformations. To apply LDDMM, we performed voxel-wise diffeomorphic registration on 1~mm isotropic MR images, and the template mesh was then adjusted based on the optical flow derived from the deformation field. ShapeWorks, in contrast, represents a correspondence-based method that identifies point correspondences across a population of 3D point clouds derived from individual shapes.



\begin{figure*}[!t]
\includegraphics[width=\textwidth]{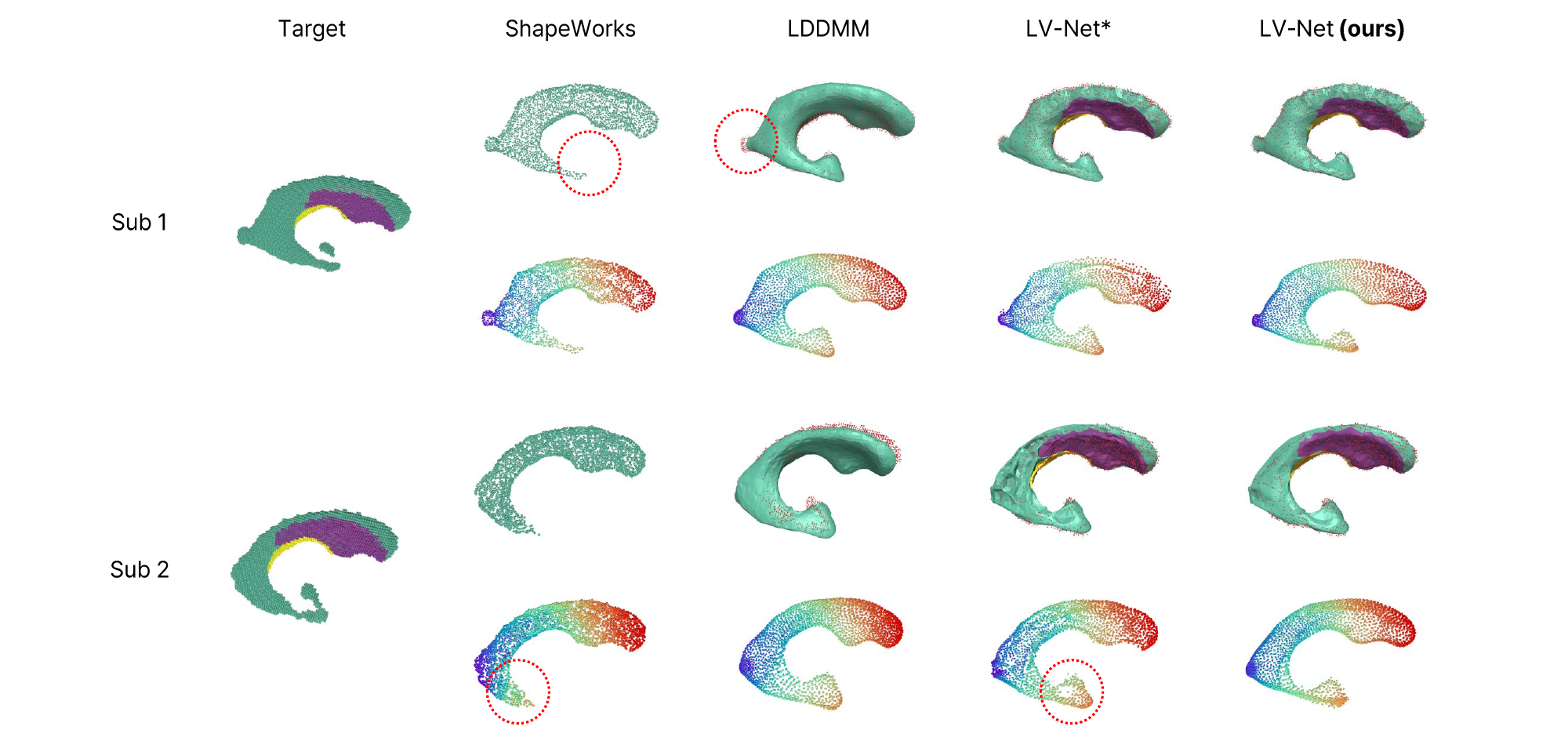}\centering
\caption{Examples of optimization results according to the shape modeling methods. Red points are target shapes on the first row of each subject. The colors of spectral particles indicate correspondence for each method. LV-Net* indicates that when using only the LV template mesh $M_l$ along with periventricular information.} \label{fig:all_optim}
\end{figure*}


\begin{table}[!t]
\caption{Shape alignment accuracy for normal and demented subjects in the OASIS, ADNI, and AIBL dataset, and ASSD for each subcortical region from the anatomy-aware template mesh.}
\label{tab:acc}
\centering
\fontsize{8pt}{10pt}\selectfont 
\begin{tabular}{c|c|c|c}
\toprule
 Method   & DSC & ASSD (mm) &HD95 (mm)  \\ \hline
  SW &-&0.952&2.681 \\
LDDMM     &0.775&1.469&3.495  \\
 Ours  &\textbf{0.893}&\textbf{0.715}&\textbf{2.064} \\
 \bottomrule
\end{tabular}
\end{table}

\subsection{Evaluation on LV shape modeling.}
To evaluate in a multifaceted manner, we report both LV shape alignment accuracy and the corresponding visualization in \autoref{tab:acc} and \autoref{fig:all_optim}.
As shown in the table, LV-Net outperforms all methods across Dice score (DSC), averaged symmetric surface distance (ASSD), and Housdroff distance at 95 percentile (HD95), while consistently producing smooth surfaces with uniformly distributed points.
In contrast, the traditional methods show inferior performances in the alignment scores with partially reconstructed LV shapes. 
Especially, LDDMM 
fails to reach the target point cloud with the high ASSD and HD values. ShapeWorks
suffers to model the inferior LV, causing inaccurate correspondences across subjects, and its lack of vertex connectivity results in locally disordered points, as seen in the spectral view.
Although LV-Net*, which uses only the LV part of the template mesh along with periventricular information, exhibits well-deformed shapes to the target point clouds, the qualitative results exhibit instability with non-uniformly distributed points
as well as unrealistic LV anatomy such as skewed and folded structures.
These artifacts and incomplete shapes not only degrade anatomical fidelity but also render the resulting shapes unreliable for further analysis or clinical interpretation. This underscores the importance of using a joint LV-hippocampus template mesh which compensates for incomplete segmentation in the inferior part of the LV.

\begin{figure}[]
\centering
\includegraphics[width=\linewidth]{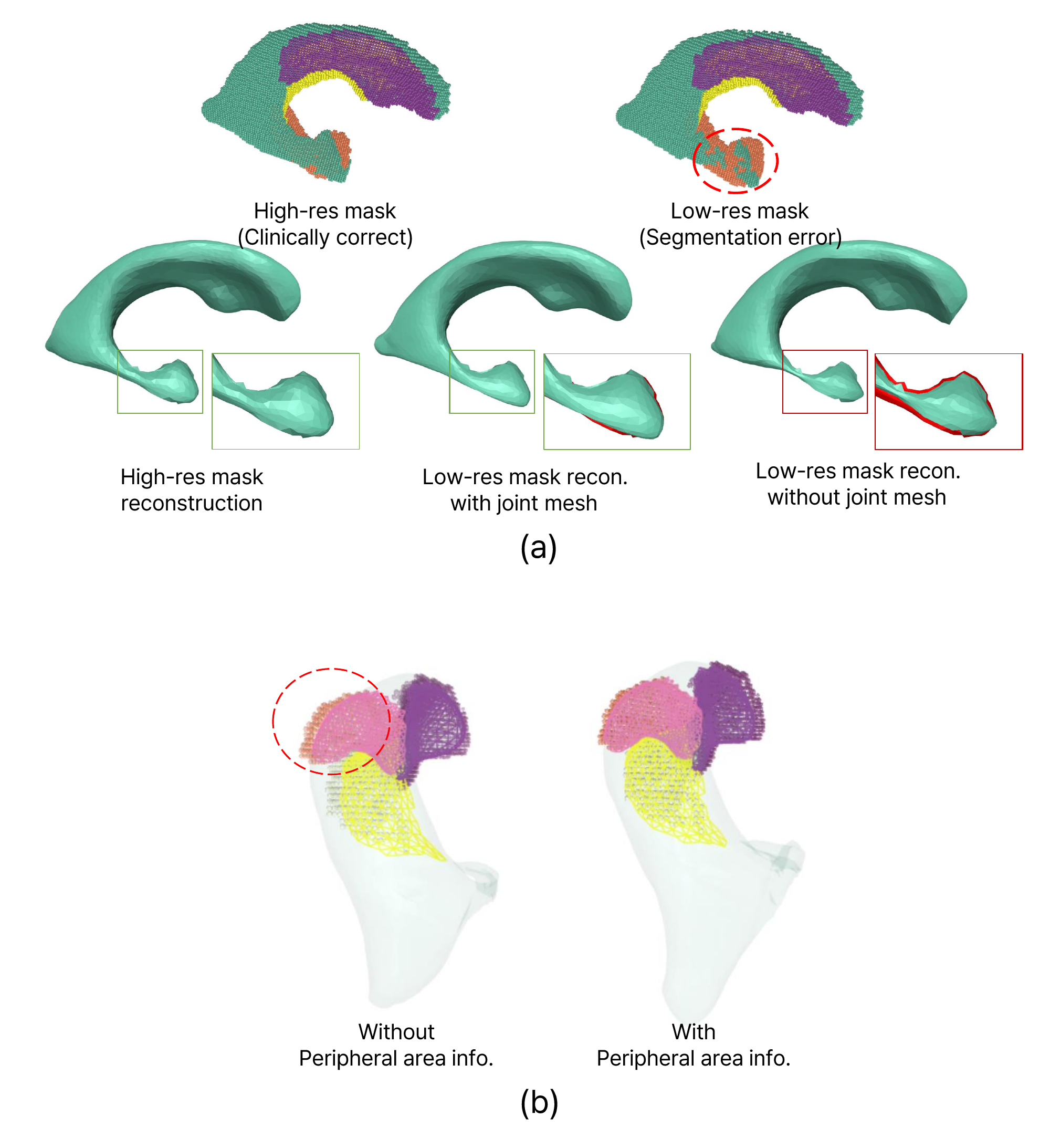}
\caption{(a) Error correction by joint template mesh-based LV reconstruction. The inferior LV is accurately restored to match the high-resolution shape. (b) Guiding template vertices to corresponding subcortical areas using peripheral region information yields more reasonable results.}
\label{fig:abl}
\end{figure}
\subsection{Anatomy correction with the template mesh.}
Our LV-Net incorporates the anatomical knowledge of the LV-hippocampus by using the joint template mesh to compensate for segmentation errors. 
LV-Net*, which solely uses LV template mesh, the absence of guidance due to missing LV masks induces skewed mesh artifacts (i.e. non-uniformly distributed points) in the inferior LV.
On the other hand, our LV-Net maintains uniformly distributed points in the inferior LV while accurately representing the target point clouds.

To further validate the error correction capability, we conducted an ablation study by simulating segmentation errors through downsampling high-resolution masks. We then reconstructed the low-resolution masks with and without the joint template mesh and compared them to the high-resolution reconstruction. As illustrated in \autoref{fig:abl}(a), the joint template mesh effectively restores the inferior LV to a shape nearly identical to the high-resolution reconstruction, demonstrating its anatomical correction capability.

To ensure precise point correspondence across subjects, we leverage peripheral area information in modeling LV shapes.
As depicted in \autoref{fig:abl}(b), by utilizing guidance for each subcortical region with the $\mathcal{L}_{dist}^\text{peri}$ in \autoref{equ:dist},LV-Net succesfully guide template mesh vertices to exhibit conspicuously coherent deformation within corresponding peripheral regions with the guidance of the anatomy-aware template mesh.

\begin{figure*}[!t]
\includegraphics[width=0.95\textwidth]{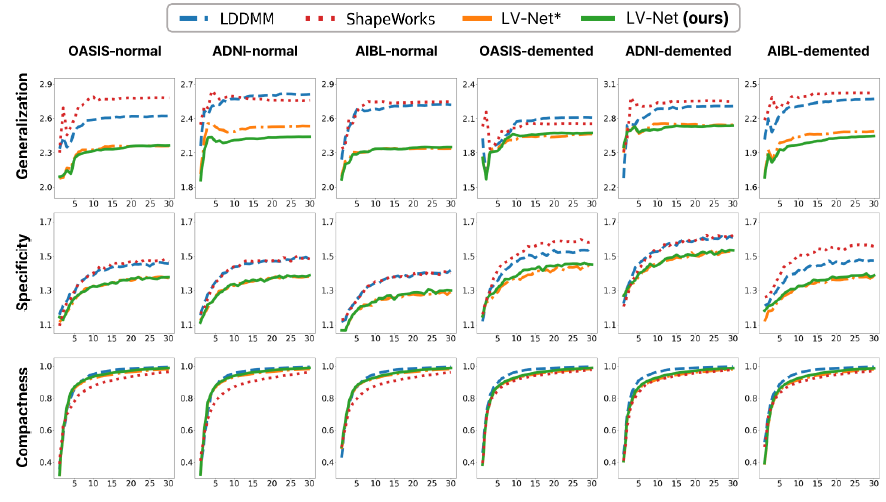}\centering
\caption{Quantitative plots of generalization, specificity and compactness on normal and demented groups of OASIS, ADNI, and AIBL datasets with 50 subjects each.} \label{fig:PCA_metric}
\end{figure*}

\begin{figure*}[!t]
\includegraphics[width=\textwidth]{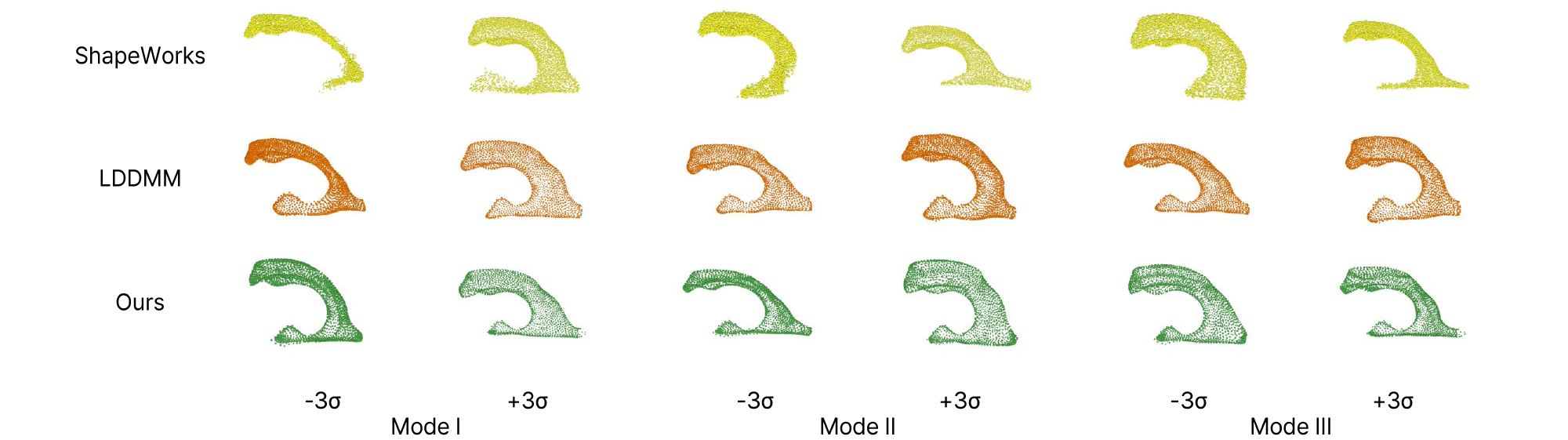}\centering
\caption{Visualization of the first 3 PCA components of the learned LV shape spaces. We modulated the shape in each dimension using $3\sigma$ and -$3\sigma$ variations.}\label{fig:pca}
\end{figure*}


\subsection{Evaluation on statistical shape modeling quality.}
Since the ultimate goal of LV shape modeling is to analyze group-wise shape statistics, we examine the statistical shape modeling quality using three standard metrics: generalization, specificity, and compactness. Through this assessment, we verify the LV-Net capacity for capturing the high LV shape variances.
As illustrated in \autoref{fig:PCA_metric}, LV-Net outperforms LDDMM and ShapeWorks by a large margin and achieves higher or comparable results to LV-Net* across all datasets and metrics. In \autoref{fig:pca}, the PCA results demonstrate that our LV-Net capable of capturing a broader range of LV shapes with large morphological variations.
Otherwise, with traditional methods, LDDMM exhibits lower variance, and ShapeWorks includes anatomically incorrect shapes despite showing larger variance. Considering the qualitative and quantitative results with the PCA visualization, our LV-Net is beneficial to reconstruct individual LV shapes.

\section{Application of LV-Net to Alzheimer's Disease Shape Analysis}

\begin{figure}[]
\centering
\includegraphics[width=\linewidth]{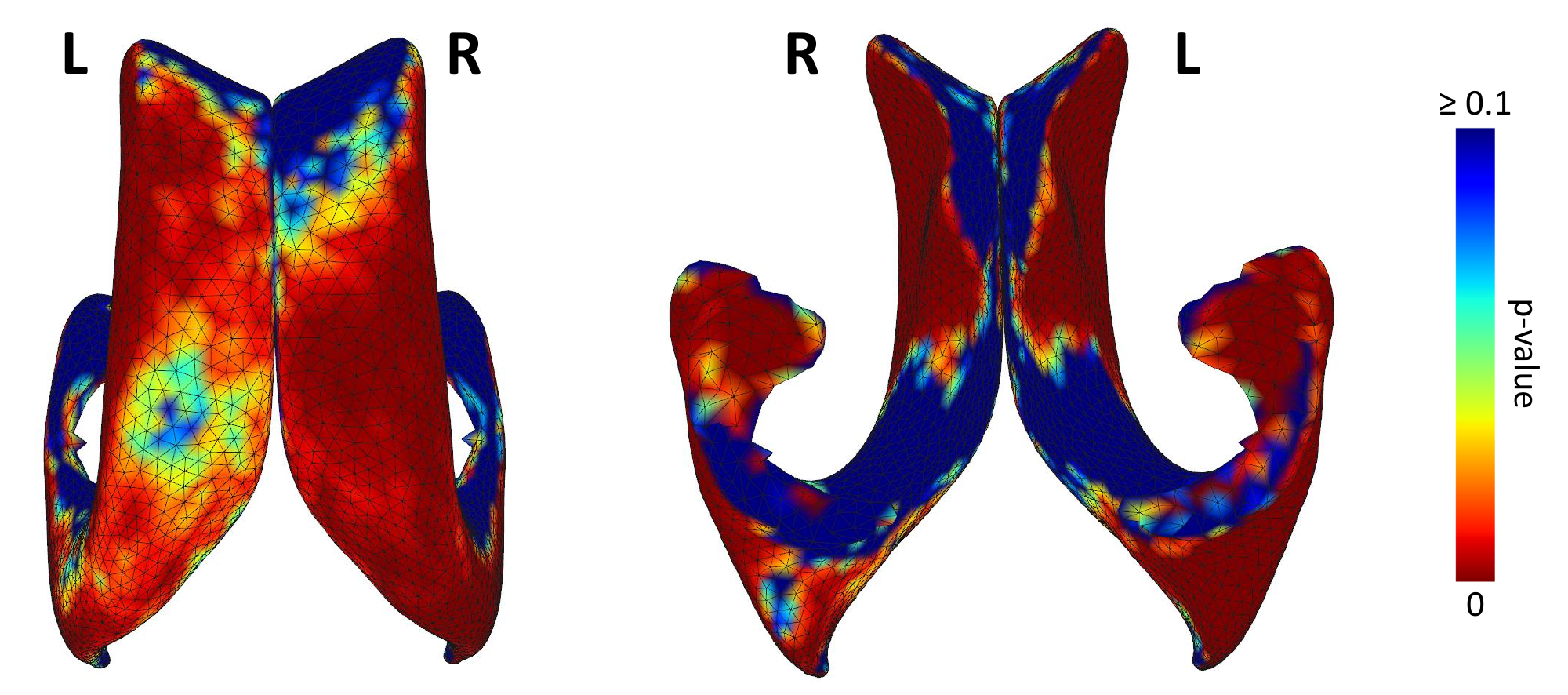}
\caption{Visualization of the lateral ventricle indicating subregions where vertex-wise shape distributions significantly differ between Alzheimer's disease subjects and cognitively normal controls. (superior and inferior views)}
\label{fig:pvalue}
\end{figure}

In addition to introducing LV-Net and presenting its evaluations, we demonstrate its clinical application in this section. We apply the LV-Net to identify subregions of the lateral ventricle (LV) that are significantly associated with Alzheimer's Disease (AD) compared to normal controls. For this analysis, we modeled 299 LV shapes from cognitively normal subjects (mean age: 76.5 ± 5.58 years) and 179 LV shapes from AD patients (mean age: 74.6 ± 5.58 years). We assess shape difference using Wilcoxon signed rank test since the non-normal distribution of vertex data is confirmed by the Shapiro-Wilk test~\cite{shaphiro1965analysis_normality}. 

We visualized the regions where vertex-wise shape distributions significantly differed between the two groups ($p\leq0.1$), as shown in \autoref{fig:pvalue}. Consistent with prior studies, the hippocampal region shows a strong association with Alzheimer's disease (AD). In addition, the portions of the LV adjacent to the thalamus and to parts of the caudate also exhibit significant associations with AD. These findings align with volumetric morphology analyses in subcortical areas, which report large mean volume differences in the hippocampus and thalamus between AD and cognitively normal subjects, while the caudate shows relatively smaller differences~\citep{watson2016subcortical}.

\section{Conclusion}
In this work, we introduced LV-Net, an automated framework for reconstructing lateral ventricular (LV) shapes in mesh representation from brain MRIs. To enhance anatomical accuracy, we proposed an anatomy-aware joint template mesh, which enables robust shape reconstruction despite substantial inter-subject variability and segmentation errors. Through comprehensive experiments, we quantitatively validated the effectiveness of our framework in both shape alignment and statistical shape modeling.
Furthermore, we demonstrated that the anatomy-aware joint template mesh improves anatomical point correspondence and effectively corrects segmentation artifacts. A case study on Alzheimer’s disease illustrated the utility of LV-Net in capturing disease-related LV shape characteristics. Our LV-Net not only improves global LV shape reconstruction but also facilitates localized analysis of surrounding subcortical structures, providing a solid foundation for advancing
neurological disease research.
\section*{Acknowledgments}
This work was supported by the National Research Foundation of Korea (NRF) grant funded by the Korea government (MSIT) (RS-2024-00508681, Establishment of Korea-UK preclinical/clinical joint research center to develop diagnosis and treatment strategy for neurodegenerative diseases).

Data collection and sharing for this project was funded by the Alzheimer's Disease Neuroimaging Initiative (ADNI) (National Institutes of Health Grant U01 AG024904) and DOD ADNI (Department of Defense award number W81XW H-12-2-0012). ADNI is funded by the National Institute on Aging, the National Institute of Biomedical Imaging and Bioengineering, and through generous contributions from the following: AbbVie, Alzheimer’s Association; Alzheimer’s Drug Discovery Foundation; Araclon Biotech; BioClinica, Inc.; Biogen; Bristol-Myers Squibb Company; CereSpir, Inc.; Cogstate; Eisai Inc.; Elan Pharmaceuticals, Inc.; Eli Lilly and Company; EuroImmun; F. Hoffmann-La Roche Ltd and its affiliated company Genentech, Inc.; Fujirebio; GE Healthcare; IXICO Ltd.; Janssen Alzheimer Immunotherapy Research \& Development, LLC.; Johnson \& Johnson Pharmaceutical Research \& Development LLC.; Lumosity; Lundbeck; Merck \& Co., Inc.; Meso Scale Diagnostics, LLC.;  NeuroRx Research; Neurotrack Technologies; Novartis Pharmaceuticals Corporation; Pfizer Inc.; Piramal Imaging; Servier; Takeda Pharmaceutical Company; and Transition Therapeutics. The Canadian Institutes of Health Research is providing funds to support ADNI clinical sites in Canada. Private sector contributions are facilitated by the Foundation for the National Institutes of Health (www.fnih.org). The grantee organization is the Northern California Institute for Research and Education, and the study is coordinated by the Alzheimer’s Therapeutic Research Institute at the University of Southern California. ADNI data are disseminated by the Laboratory for Neuro Imaging at the University of Southern California. 

\printcredits

\bibliographystyle{cas-model2-names}

\bibliography{cas-refs}

\end{document}